\documentclass{tlp}
\title{Actual causation in CP-logic}
\author{Joost Vennekens\\ Lessius Mechelen (Campus De Nayer), Belgium}
\newcommand{\true}{{\bf t}}
\newcommand{\false}{{\bf f}}

\usepackage[colorlinks=false]{hyperref}  
\hypersetup{linkbordercolor={1 1 1},
citebordercolor={1 1 1},
urlbordercolor={1 1 1}
}
\usepackage{verbatim}
\usepackage[all]{xypic}
\usepackage{pgf, pgfnodes, pgfarrows}
\usepackage[sumlimits]{amsmath}
\usepackage{amssymb}
\usepackage{amstext}
\usepackage{amscd}

\newtheorem{definition}{Definition}

\newcommand{\curly}[1]{\mathcal{#1}}

\newcommand{\ci}{\curly{I}}
\newcommand{\ce}{\curly{E}}
\newcommand{\cu}{\curly{U}}

\newcommand{\pts}{\curly{T}}

\begin{document}
\maketitle

\begin{abstract}
Given a causal model of some domain and a particular story that has taken place in this domain, the problem of actual causation is deciding which of the possible causes for some effect actually caused it. One of the most influential approaches to this problem has been  developed by Halpern and Pearl in the context of structural models.  In this paper, I argue that this is actually not the best setting for studying this problem.  As an alternative, I offer the probabilistic logic programming language of CP-logic.  Unlike structural models, CP-logic incorporates the deviant/default distinction that is generally considered an important aspect of actual causation, and it has an explicitly dynamic semantics, which helps to formalize the stories that serve as input to an actual causation problem. 

\end{abstract}

\section{Introduction}
Actual causation has puzzled philosophers since at least the work by \citeN{lewis73}.  One way of phrasing the problem is as follows: suppose we know the causal laws that govern some domain, and that we then observe a story that takes place in this domain; when should we now say that, in this particular story, one thing actually caused another? Recent work by \citeN{halpernpearl05a} has also garnered interest in this topic in the AI community.  Their account (which I will refer to as {\em HP}) constructs a formal definition that tries to capture this intuition in the context of structural models \cite{pearl:book}.  To be more concrete, it  defines when  some random variables $\vec{X}$ of the structural model having values $\vec{x}$ can be counted as an actual cause for $\vec{Y} = \vec{y}$.

In previous work, I have tried to show that the knowledge representation properties of Pearl's structural models can be improved by borrowing representations and techniques from logic programming.  In particular, \citeN{vennekens04} introduced the probabilistic logic programming language of Logic Programming with Annotated Disjunctions, for which Riguzzi (2008; 2010) \nocite{riguzzi08,riguzzi10}
implemented SLD and SLG based resolution algorithms.  Further analysis of this language has lead to a reformulation of its semantics, called {\em CP-logic}, which attempts to clarify its causal aspects and examine its relation to  Pearl's work \cite{vennekens09}.  A more recent paper \cite{vennekens:jelia} showed that Pearl's analysis of interventions and counterfactuals in the context of  structural models can be elegantly redone in the context of CP-logic, yielding better results for a number of examples, most notably when cyclic causalities are involved.

The goal of this paper is to examine the notion of actual causation in the context of CP-logic.  Section \ref{sec:mot} will start with some motivation, by explaining a few of the differences between structural models and CP-logic, and offering some hand-waving arguments for why CP-logic might offer a more appropriate setting for the study of actual causation.  The semantics of CP-logic is briefly recalled in Section \ref{sec:prel}.  Then, Section \ref{sec:hp} dives into the details of actual causation by discussing the HP definition, while Section \ref{sec:ac} gives my own account.  The traditional way of testing such a definition is to run through a number of  ``tricky'' examples and checking whether the  obtained answers are intuitively plausible.  In Section \ref{sec:ex}, I will follow suit.  Finally, Section \ref{sec:imp} briefly comments on a  naive implementation in Prolog that can be downloaded  to play with my definitions. 

\section{Motivation: structural models and CP-logic}\label{sec:mot}

A structural model is based on a set of random variables (RVs).  Each RV has an associated domain of possible values.  The simplest case are Boolean RVs, which have $\{\true,\false\}$ as their domain, and can therefore be thought of as propositional symbols or ground atoms. Boolean RVs suffice for typical examples of actual causation,  so I will from now on restrict attention to just these.  

A structural model then consists of a set of equations $X := f(\vec{Y})$,
which define the value of the RV $X$ in terms of the RVs $\vec{Y}$ by a Boolean function $f$.  The RVs that appear in the left-hand side of such an equation are called {\em endogenous}, and the other ones {\em exogenous}. Typically, these sets of equations are assumed to be acyclic. Their meaning is formalized by the obvious possible world semantics: each assignment of values to the RVs that satisfies all of the equations is a possible world. An acyclic set of equations has the useful property that an assignment of values to the exogenous RVs uniquely determines a single possible world.

Pearl uses structural models to represent causal relations: each equation $X := f(\vec{y})$ is taken to mean that the causes for $X$ taking value $x$ are all assignments $\vec{Y}=\vec{y}$ for which $f(\vec{y})=x$.  This representation is used by Pearl to great effect, studying interventions, counterfactuals, and of course also actual causation.

Despite its successes, however, there is something peculiar about the structural model representation of causal relations:  it does not take into account their dynamic nature. Suppose, for instance, that you make the causal claim that  dropping a glass causes it to break.  If I don't believe you, I might challenge you to prove your claim. How would you do this? Presumably, you would first hold out an unbroken glass.  Then, you would drop it, so that I could watch it fall, hit the floor, and break.  In other words, you would show me a  transition from a state of the world in which the glass is whole to a one in which it is broken.  If you can convince me that it was indeed your dropping the glass that initiated this transition, then you have proven your causal claim. 

What this little thought experiment shows is that the idea of a transition from one state of the world to the next is inherently part of the way in which we interpret causal statements.  However, structural models have nothing to do with such transitions.  For instance, the causal claim about glasses breaking would just be represented by an equation $Break := Drop$, which determines two possible worlds: $\{ Break = \true, Drop =\true\}$ and $\{ Break = \false, Drop =\false\}$.  In this sense, structural models make complete abstraction of the dynamic aspects of causality, until all that remains is a static picture of how the values of different random variables can be defined in terms of each other.

Other approaches to causality do not share this static worldview. For instance,  \citeN{shafer:book} gives an explicitly dynamic  account of causation.  He represents causal systems by means of probability trees, in which edges represent transitions between states of the world.  For example, in the following picture, the edge going from $N_1$ to $N_2$ represents a transition from a state  in which Joe hasn't yet taken a swing at the ball to one in which he has and missed:
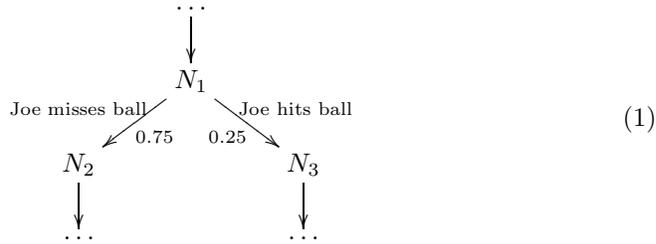
\begin{equation}
\begin{split}\label{transition}
\xymatrixrowsep{0.6cm}
\xymatrix{
&\ldots\ar[d]\\
& N_1 \ar[ld]_{\text{Joe misses ball}}^{0.75} \ar[rd]^{\text{Joe hits ball}}_{0.25}\\
N_2 \ar[d]& & N_3\ar[d]\\
\ldots && \ldots
}
\end{split}
\end{equation}
 The edge $(N_1,N_3)$ represents a transition of the same  state $N_1$ to a state where Joe has hit the ball.  Together, these two edges represent a non-deterministic event\footnote{Note that I do not use  the term ``event'' in its probability-theoretical meaning of ``a set of possible outcomes'', but rather in the common-sense meaning of ``a thing that happens''. }, namely that of Joe's taking a swing at the ball, which may result in one of these two outcomes. 
   The edges are labeled with the probabilities of the outcomes: the probability of Joe's swing missing is $0.75$ and that of it hitting is $0.25$. 

This paper will use {\em CP-logic} as its formal language, which is essentially just a modular, syntactic representation for such Shaferian probability trees.  A theory in CP-logic represents the causal structure of a domain by means of a set of {\em causal probabilistic laws} (CP-laws, for short).   Each such CP-law is a blue-print for a class of non-deterministic events.  For instance, the following CP-law:
\[ \forall p,b\ (Hit(p,b):0.25) \lor (Miss(p,b):0.75) \leftarrow Swing(p,b)\]
states that, for every player $p$ and ball $b$, player $p$'s taking at swing at ball $b$ causes a non-deterministic event, which has as one possible outcome that $p$ hits $b$ (and this happens with probability $0.25$), and as its other possible outcome that $p$ misses $b$ (which happens with probability  $0.75$).   

If $p$ and $b$ are instantiated to, respectively, a particular player and a particular ball, say Joe and the twelfth pitch, the we obtain a description of one particular event (that may or may not happen, depending on whether Joe decides to swing):
\begin{equation*} (Hit(Joe,12):0.25) \lor (Miss(Joe,12):0.75) \leftarrow Swing(Joe,12). \label{joeshit}
\end{equation*}
This instantiated CP-law can be seen as a textual representation of picture \eqref{transition}, provided that, of course, node $N_1$ represents a state of the domain where Joe has decided to take a swing at this particular pitch. In this way, each instantiation of a CP-law describes a piece of probability tree.  As will be explained in more detail later, an entire CP-theory describes a class of probability trees, each of which can be constructed by putting these small pieces together.

Unlike structural models, the formal semantics of CP-logic therefore does provide mathematical objects that represent transitions between states of the world.  As argued by \citeN{stonesoup}, these transitions are important for a study of actual causation.  Indeed, when the goal is to figure out what caused what in a given {\em story}, it is obviously convenient to have a language whose formal semantics already offers objects that correspond in  a natural way to stories.  A branch of a Shaferian probability tree is precisely such an object, because, like a story, it is a description of a sequence of events that change the state of the world.

There is also a second argument in favour of CP-logic. The goal of actual causation is to explain why things happened.  Typically, though, not everything is in need of explanation.  A detective solving a murder case, for instance, will be interested in why the victim is dead, but he won't care about why he was ever alive in the first place.  The detective's  causal model will therefore list causes for dying (poison, gun shot, \ldots), but not for living (sexual intercourse, IVF, \ldots).   In more technical terms, the detective considers living to be the {\em default} state of a person, and he is only interested in {\em deviations} from this default.   Many authors, such as \citeN{hall04} or \citeN{hitchcock07}, have argued that actual causation should be studied under the assumption that each RV has such a default state.

Structural models make no distinction between the different values of a RV.  Consequently,  a RV $Alive$ with values $yes$ and $no$, and a RV $Dead$ with values $no$ and $yes$ are completely interchangeable.  In CP-logic, this is not the case.  Here, each RV (i.e., ground atom) has {\false} as its default value.  This means that the mere existence of the atom $Alive(Adam)$ implies that the default condition is for $Adam$ to be dead, and that he can only come to life when there is a sufficient cause for this.  By contrast, the detective's theories will  contain atoms such as $Dead(Adam)$, indicating that living is the default and death is in need  of causal explanation.  In the probability trees generated by a CP-theory, an atom  always starts out at its default value, and only deviates from this when it has sufficient cause to do so.

\section{Reminder: formal semantics of CP-logic}\label{sec:prel}

Lacking space for a full review of CP-logic, I will only summarize the main ideas and refer to \cite{vennekens09} for details.  The general form of a CP-law is:
\begin{equation} \forall\vec{x}\ (A_1:\alpha_1)\lor\cdots\lor (A_n:\alpha_n)
\leftarrow \phi.\label{cplaw1}
\end{equation}
Here,  $\phi$ is a first-order formula and the $A_i$ are atoms, such that  the tuple of variables $\vec{x}$ contains all free variables in $\phi$ and the $A_i$.  The $\alpha_i$ are non-zero probabilities with $\sum\alpha_i \leq 1$.   Such a CP-law expresses that $\phi$ causes some (implicit) non-deterministic event, of which  each $A_i$ is a possible outcome with probability $\alpha_i$.   If $\sum_i \alpha_i = 1$, then at least one of the possible effects $A_i$ must result if the event caused by $\phi$ happens; otherwise, it is also possible that the event happens without any (visible) effect on the state of the world.  For the purpose of this paper, the propositional fragment of CP-logic suffices, so I will from now on restrict attention to CP-laws in which the tuple of variables $\vec{x}$ is empty. 

For a CP-law $r$, we refer to $\phi$ as the {\em body} of $r$, and to the sequence $(A_i,\alpha_i)_{i = 1}^{n}$ as the {\em head} of $r$.    We denote these objects as $body(r)$ and $head(r)$, respectively, and also write $head_{At}(r)$ for the set of all $A_i$ for which there exists an $\alpha_i$ such that $(A_i,\alpha_i) \in head(r)$. For CP-laws that  are  vacuously caused, $body(r)$ may  be omitted.  If a CP-law has a deterministic effect, i.e., it is of the form $(A:1)\leftarrow\phi$, it is also written simply as $A\leftarrow\phi$.
 
A {\em CP-theory} is a finite set of CP-laws. Such a CP-theory describes the non-deterministic evolution of a domain, which is formally represented by a Shaferian probability tree.  Initially, all RVs of this domain (i.e., all ground atoms) are in their default state.  This means that we can describe the initial state of the domain, which corresponds to the root of the probability tree, by the interpretation that assigns {\false} to each of them.  We then extend this root by picking a CP-law $r$ whose precondition $body(r)$ is satisfied according to this interpretation and creating a child node for each pair $(h_i: \alpha_i)$ in $head(r)$.  The edge to child $i$ is labeled with the probability $\alpha_i$ and the corresponding new state of the domain is constructed from the previous state by switching $h_i$ to its deviant state {\true}.   The CP-law $r$ has now happened, and will not happen again.

We repeat this process of adding children to one of the leaf nodes of the current tree, until this is no longer possible, i.e., until for all leaves $l$ of the current tree it is the case that all rules $r$ that have not yet happened in $l$ have a precondition $body(r)$ that is false in $l$.  The resulting trees are called the {\em execution models} of the CP-theory.  For a node $s$ of the tree, I denote by $\ci(s)$ the interpretation that corresponds to the state of the world at that node, and, if $s$ is not a leaf, by $\ce(s)$ the CP-law that was used to create the children of this node.  

The construction of execution models is quite non-deterministic, in the sense that in any particular node of the tree, there can be many CP-laws that may be used to extend it.  The question is now whether each of these trees actually reflects a sensible way in which a domain described by the CP-theory might evolve.  The answer is a qualified ``yes'', and depends on precisely how we choose to interpret negation appearing in the body of a CP-law.  Consider the following example:
\[
 (Shatters: 0.9) \leftarrow \lnot DecidesNotToThrow(Suzy).
\]
We could take the body of this CP-law to mean that this transition may happen in any state where $DecidesNotToThrow$ is still at its default state \false, such as, by definition, the initial state in which Suzy has no yet made up her mind about throwing. Taking this view, every probability tree constructed according to the above principles can indeed be seen as a sensible description of how the domain might evolve.   However, this is not very useful.  As argued by \citeN{vennekens09}, it is more interesting to read negation in a slightly different way, namely, as not just saying that  $DecidesNotToThrow$ is still at its default value in the current state, but that it can actually {\em never} deviate any more.  In other words, according to this reading, the above CP-law will only be applicable {\em after} Suzy has decided that she will not refuse the throw.    This idea is formalized in the semantics of CP-logic by means of a construction similar to the Gelfond-Lifschitz reduct.  We use this to compute, for each state $s$, an overestimate $\cu(s)\supseteq \ci(s)$ of all atoms that can still be caused in this $s$.  Only if an atom $a$ does not belong to $\cu(s)$, do we then say that $\lnot a$ holds in $s$.  If there are no loops containing double negation (i.e., some $\lnot P$ causing $Q$ and $\lnot Q$ causing $P$), then it is the case that, in any branch of a probability tree, each CP-law must either happen at some point, or else become impossible.  \citeN{vennekens09}  showed that there is a close connection between the resulting semantics and the well-founded model construction for a logic program.   

Each probability tree $\pts$ defines, in the obvious way, a  probability distribution $\pi_\pts$ over its leaves.   For an execution model $\pts$ of a CP-theory $C$, this distribution $\pi_\pts$ induces a probabilistic possible world semantics: the probability $\pi_\pts(S)$ of an interpretation $S$ is $\sum_{\ci(l) = S} \pi_\pts(l)$, where the sum is taken over leaves $l$ of $\pts$.  \citeN{vennekens09} showed that each execution model $\pts$ of a CP-theory $C$ defines the same possible world semantics $\pi_\pts$.  For instance, the two trees shown on page \pageref{trees} are execution models of the same theory and, even though they are not isomorphic, they  both define the same $\pi_\pts$.  In this way, each CP-theory $C$ defines a unique probability distribution, which is denoted as $\pi_C$.  The probability of a formula $\phi$ can then be defined as  $\pi_C(\phi) = \sum_{S \models \phi}\pi_C(S)$.  

The fact that $\pi_C$ does not depend on the choice of any particular execution model $\pts$  may help to explain why structural models choose to ignore the dynamic aspects of causality in the first place.  Indeed, this result shows precisely that, for applications which only care about properties of the final state that the domain will eventually reach, the details of how this final state came about can be safely ignored.  As I attempt to show in this paper, though, actual causation is {\em not} such an application.

Like structural models, CP-logic also makes a distinction between exogenous and endogenous random variables.  With $X$ the set of all exogenous atoms, the semantics of a CP-theory now becomes relative to an interpretation $I$ for these atoms.  In particular, an execution model for $C$ {\em given $I$} is defined as a execution model that starts not from a root in which {\em all} atoms are {\false}, but instead starts with only the endogenous atoms being {\false} and the exogenous atoms being interpreted by $I$.  \citeN{vennekens09} have shown that for each interpretation $I$ for the exogenous predicates of a CP-theory $C$, all execution models $\pts$ given $I$ define the same probability distribution $\pi_\pts$, which is denoted as $\pi_C^I$.  

\section{Actual causation in HP}\label{sec:hp}

This section briefly recalls the HP account.  Their paper starts with this example:
\begin{quote}
Suppose that two arsonists drop lit matches in different parts of a dry forest, and that both cause trees to start burning, until the entire forest burns down. Both matches are necessary to burn down the forest; with only one match, the fire would die down.
\end{quote}
It is clear that both arsonists are an actual cause of the forest burning down.  HP reach this conclusion as follows.  To represent the causal structure of the example, they use a structural model consisting of a single equation:
\begin{equation}
Burn := Match_1 \land Match_2. \label{matches1}
\end{equation}
The particular story under consideration is then represented by the following assignment of values to the exogenous RVs: $\{  Match_1 = \true, Match_2 = \true \}$.   This of course also uniquely determines the values of the endogenous RVs: $Burn =\true$.

The HP  definition is now reproduced below.  In it, $M$ is a structural model with endogenous RVs $\curly{V}$, $\vec{u}$ an assignment of values to the exogenous RVs, $\vec{X}$ a tuple of endogenous RVs, and $\phi$ a Boolean formula in the RVs.   The notation $(M,\vec{u}) \models [\vec{X}\leftarrow\vec{x}] \phi$ means that $\phi$ holds in $(M,\vec{u})$ after the intervention of assigning $\vec{x}$ to $\vec{X}$ is performed, i.e., each $X_i \in \vec{X}$ has its defining equation removed from $M$ and replaced by $X_i := x_i$. 

\begin{definition}[HP account of actual causation]\label{HP}
$\vec{X}=\vec{x}$ is an actual cause of $\phi$ in $(M, \vec{u})$ if the following three conditions hold.
\begin{itemize}
\item[AC1.] $(M, \vec{u}) \models (\vec{X}⃗ =\vec{x}) \land \phi.$ (That is, both $\vec{X} =\vec{x}$ and $\phi$ are true in the actual world.)
\item[AC2.] There exists a partition $(\vec{Z},\vec{W})$ of $\mathcal{V}$ with $\vec{X}\subseteq \vec{Z}$ and some setting $(⃗\vec{x}',\vec{w}')$ of the variables in $(\vec{X},\vec{W})$ such that if $(M, \vec{u}) \models Z = z^{*}$ for all $Z \in \vec{Z}$, then both of the following conditions hold:
\begin{itemize}
\item[(a)] $(M,\vec{u}) \models [\vec{X}\leftarrow \vec{x}', \vec{W}\leftarrow \vec{w}']\lnot\phi$. In words, changing $(\vec{X},\vec{W})$ from $(\vec{x},\vec{w})$ to $(\vec{x}',\vec{w}')$ changes $\phi$ from true to false.
\item [(b)] $(M,\vec{u}) \models [\vec{X}\leftarrow \vec{x}, \vec{W}' \leftarrow \vec{w}',\vec{Z}'\leftarrow\vec{z}^*]\phi$ for all subsets $\vec{W}'$ of $\vec{W}$ and all subsets $\vec{Z}'$ of $\vec{Z}$. In words, setting any subset of variables in $\vec{W}$ to their values in $\vec{w}'$ should have no effect on $\phi$, as long as $\vec{X}$ is kept at its current value $\vec{x}$, even if all the variables in an arbitrary subset of $\vec{Z}$ are set to their original values in the context $\vec{u}$.
\end{itemize}
\item[AC3.]  $\vec{X}$ is minimal; no subset of $\vec{X}$ satisfies conditions AC1 and AC2. Minimality ensures that only those elements of the conjunction $\vec{X} = \vec{x}$ that are essential for changing $\phi$ in AC2(a) are considered part of a cause.
\end{itemize}
\end{definition}

With $\vec{X} = (Match_1)$, $\vec{Z} = (Match_1, Burn)$ and $\phi = Burn$, this definition provides the result that $Match_1$ actually caused $Fire$, since if we change $X$ to $\false$, while leaving $\vec{W} = (Match_2)$ at its original value (this trivially satisfies AC2(b)), we obtain $\lnot Burn$ as required by AC2(a).  In other words, in this example, we get actual causation from a simple counterfactual dependency: if it hadn't been for $Match_1$, the forest wouldn't have burned down.

HP also consider a disjunctive variant of this example, where a single match already suffices to burn down the forest ($Burn := Match_1 \lor Match_2$).  This causes the straightforward counterfactual criterion to fail, since stopping only one of the arsonists does not stop the forest burning down.  This motivates the additional machinery of the  above definition.   By considering the context in which $Match_2 = \false$, that is $\vec{W} = (Match_2)$ and $w' = (\false)$,  we can re-establish the counterfactual dependency of $Burn$ on $\vec{X} = (Match_1)$.

\section{Actual causation in CP-logic}\label{sec:ac}

As we have seen, a question of actual causation can only be asked in the presence of two pieces of information: a causal model of a domain (the $M$ of Definition \ref{HP})  and a story that takes place in this domain (the $\vec{u}$).  My definition will of course assume that the causal model is given in the form of a CP-theory $C$.  In the context of CP-logic, the most obvious formal counterpart of a ``story'' is a branch of an execution model of $C$. Already, this allows us some more room for nuance than HP, as the following example from \citeN{hall04} shows.
\begin{quote}
Suzy and Billy might each decide to throw a rock at a bottle.  If Suzy does so, her rock shatters the bottle with probability $0.9$.  Billy's aim is slightly worse and he only hits with probability $0.8$.  
\end{quote}
This domain corresponds to the following set of CP-laws, where $Throws(Suzy)$ and $Throws(Billy)$ are exogenous:
\begin{align}  (Shatters: 0.9) &\leftarrow Throws(Suzy).\label{suzy}\\ (Shatters:0.8) &\leftarrow Throws(Billy).\label{billy}\end{align}
Assuming that  Suzy and Billy both throw, there still exist two different execution models of the theory.  Representing the states in which the bottle is broken by an empty circle, and those in which it is still whole by a full one, they look like this:\\ 
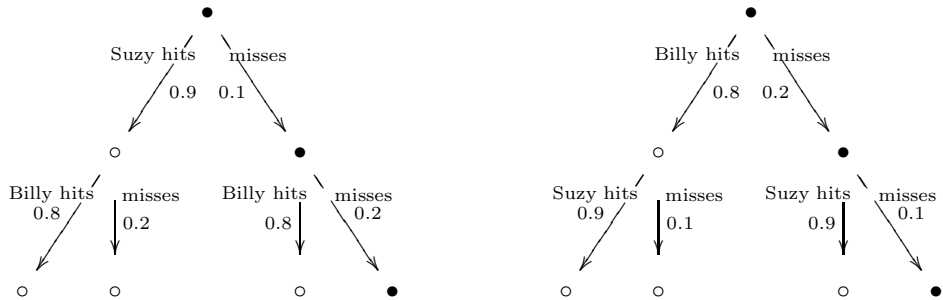
\label{trees}
\xymatrix@H=0.8cm{
&&\bullet \ar[ld]^{0.9}|(.3){\text{Suzy hits\hspace*{0.7cm}}} \ar[rd]_{0.1}|(.3){\text{\hspace*{0.6cm}misses}}\\
&\circ \ar[ld]_{0.8}|(.3){\text{Billy hits\hspace*{0.95cm}}} \ar[d]^{0.2}|(.3){\hspace*{0.95cm}\text{misses}} && \bullet \ar[d]_{0.8}|(.3){\text{Billy hits}\hspace*{0.95cm}} \ar[rd]^{0.2}|(.3){\hspace*{0.95cm}\text{misses}} \\
\circ&\circ&&\circ&\bullet
}
\hfill
\xymatrix@H=0.8cm{
&&\bullet \ar[ld]^{0.8}|(.3){\text{Billy hits\hspace*{0.7cm}}} \ar[rd]_{0.2}|(.3){\text{\hspace*{0.6cm}misses}}\\
&\circ \ar[ld]_{0.9}|(.3){\text{Suzy hits\hspace*{0.95cm}}} \ar[d]^{0.1}|(.3){\hspace*{0.95cm}\text{misses}} && \bullet \ar[d]_{0.9}|(.3){\text{Suzy hits}\hspace*{0.95cm}} \ar[rd]^{0.1}|(.3){\hspace*{0.95cm}\text{misses}} \\
\circ&\circ&&\circ&\bullet
}
\\
In the left execution model, Suzy's rock reaches the bottle before Billy's does, whereas in the right one, it is Billy's rock that gets there first.  As discussed at the end of Section \ref{sec:prel}, this difference is irrelevant if we are only interested in the final outcomes that might be reached: the probability of the bottle  shattering is $0.98$ in both models.  However, the difference becomes relevant when we want to judge actual causation.  Indeed, in the left execution model, it is possible for Suzy's rock to {\em actually} break the bottle even though Billy's also would have (in particular, this happens in the leftmost branch of the tree).  According to the execution model on the right, however, this is impossible: here, Suzy's rock can only actually break the bottle if Billy's rock fails to do so.

\citeN{hall04} goes on to consider the following story:
\begin{quote} Suzy and Billy both pick up rocks and throw them at a bottle.  Suzy's rock get there first, shattering the bottle.  Since both throws are perfectly accurate, Billy's would have shattered the bottle had it not been preempted by Suzy's throw.
\end{quote}
This story tells us precisely that we are in the leftmost branch of the left execution model above.  Hence, Suzy's rock should be the actual cause of the bottle breaking, and not Billy's.  Before showing how I reach this conclusion in the context of CP-logic, let me first remark that things are more difficult for the HP account.  Their paper first tries the following straightforward structural model:
\begin{align*} Shatters := (Throws(Suzy) \land Accurate(Suzy)) \lor (Throws(Billy) \land Accurate(Billy)).
\end{align*}
 Here,  there is no such thing as one execution in which Suzy's rock reaches the bottle first and one in which Billy's is first.  Hall's story therefore seems to say nothing more than that all five RVs are $\true$, and the phrase ``Suzy's rock gets there first'' contributes nothing.  Of course, because it is precisely this phrase that determines which rock  actually broke the bottle, this causal model does not work.

HP fix the problem by introducing two new random variables: $Hits(Billy)$ (``Billy's rock hits the (unbroken) bottle'') and $Hits(Suzy)$.  The order in which the two rocks actually reach the bottle can then be encoded {\em in the structure of the model}:
\[ Hits(Billy) := \lnot Hits(Suzy) \land Throws(Billy) \land Accurate(Billy)\]
To me, this does not seem the right way to go.  The order in which the rocks arrive is a purely contingent matter, which belongs to the details of the particular story that is being told, and {\em not} to the general causal structure of the domain.  Saying that Suzy's rock arrives before Billy's should not be placed on the same level of causal discourse as the statement that throwing rocks at bottles causes them to break.  This is not just a matter of taste, but also has practical consequences.   If we would want to know  whether Suzy's rock would still have been the actual cause of the bottle breaking if Billy's rock had gotten there first, then---in the HP account---we would not just have to look at a different story in the same domain, but we would have to change the structure of our causal model.  Such hand-tailoring of the causal model to the question under consideration is undesirable, and, as I will now show, it is not needed in CP-logic.

My definition too will be heavily based on the intuition of counterfactual dependency from a cause $C$ to an effect $E$.  Therefore, I first  formalize the following criterion: 
\begin{center} 
\hspace{\stretch{1}} \begin{minipage}{0.9\textwidth} If all events happen in the way they actually happened {\em with the exception that $C$ is somehow prevented from occurring}, then $E$ will no longer occur.
\end{minipage} \hspace{\stretch{1}} (*)
\end{center}
This requires some mathematical machinery.  First, we need to be able to fix the outcome of certain events.  For a CP-law $r$ of the form
$(A_1:\alpha_1) \lor \cdots \lor (A_n:\alpha_n)\leftarrow \phi$,
we write $r^{A_i}$ to denote the deterministic CP-law $A_i \leftarrow \phi$.  If we now have a branch $b$ that tells us what actually happened, then we can define as follows a theory that fixes the outcome of all events that happened to their actual outcome.

\begin{definition}
Let $b = (s_0,\ldots,s_n)$ be a branch of an execution model of a CP-theory $T$.  We define $T^b$ as the union of two disjoint sets $S_1$ and $S_2$, where  $S_1$ contains all CP-laws from $T$ that did not happen in branch $b$, i.e., $ S_1 =  T \setminus \{ \ce(s_i) \mid 0 \leq i < n \}$, 
and $S_2$ consists of all $r^A$ for which $r$ caused $A$ in $b$, i.e., 
\[ S_2 = \{ r^A \mid r \in T\text{ and for some } i: \ce(s_i) = r\text{ and }\ci(s_{i+1}) \setminus \ci(s_i)  = \{A\}\}.\]
\end{definition}

We also need an antonymical transformation, which prevents some $A_i$ from occurring.  For an $r$ of the same form as above, we write $r^{\lnot{A_i}}$ for:
\[  (A_1:\alpha_1) \lor \cdots \lor (A_{i-1}:\alpha_{i-1}) \lor (A_{i+1}:\alpha_{i+1}) \lor \cdots \lor (A_n:\alpha_n)\leftarrow \phi.\]
To prevent an atom $A$ entirely, it now suffices to apply this transformation to all CP-laws that might cause it.  Given a theory $T$, we therefore define $T^{\lnot A}$ as:
\[T^{\lnot A} = \{ r^{\lnot A} \mid r\in T\text{ and }A \in head_{At}(r) \}  \cup \{ r \mid r\in T\text{ and  } A \not \in head_{At}(r) \}.\]
By combining this transformation with the previous one, we can now construct a theory $(T^b)^{\lnot C}$ which corresponds precisely to the counterfactual eventuality that everything happens precisely as it did in branch $b$, with the exception that $C$ is somehow prevented from occurring.  I thus formalize the counterfactual criterion (*), by expressing that, according to this new CP-theory $(T^b)^{\lnot C}$, $E$ will not occur. 

\begin{definition} Let $b = (s_0,\ldots,s_n)$ be a branch of an execution model of a theory $T$.
For two atoms $C$ and $E$, such that both $C$ and $E$ hold in $\ci(s_n)$, we say there is a {\em counterfactual dependency} from $C$ to $E$ if $\pi_{T'}^{I'}(E) = 0$  where $T' =  (T^b)^{\lnot C}$ and, to cover the case where $C$ is exogenous, $I'$ is $\ci(s_0) \setminus\{C\}$.
\end{definition}

Here, saying that $\pi_{T'}(E) = 0$ is of course equivalent to $E$ being false in each leaf $l$ of each execution model of $T'$.

This intuition of counterfactual dependency forms the core of the concept of actual causation, but as discussed above, it is in itself not enough. The additional aspect is the idea of {\em relevance}.  A causal model might make provisions for a large number of eventualities, many of which may not have been relevant in the actual course of events.  It is typical for judgments of actual causation that truly irrelevant causal mechanisms are ignored, even when they might appear to become relevant in a counterfactual context.
 
The typical case where this intuition manifests itself is when counterfactual dependencies are masked by {\em redundant causation}: there is some back-up mechanism waiting in the wings, which will ensures that the effect happens anyway, even if we preempt its actual cause.  The example of Suzy and Billy is a good illustration of this.  The reason why we nevertheless insist  that Suzy is the actual cause of the bottle shattering is precisely a criterion of relevance: because Suzy's rock got to the bottle first, Billy's was irrelevant, so we ignore it.

\citeN{pearl:book}  tried to formalize this same intuition by means of the concept of a {\em causal beam}, which is meant to encompass precisely the relevant parts of the causal model.  However, the formal details proved hard to get right, and the refinement that eventually became part of the HP definition seems to be a  significant source of complexity, which considerably clouds the otherwise rather simple idea of counterfactual dependency.  In the explicitly dynamic context of CP-logic, something much more simple is possible.

Let us ask again why intuitions feels that Billy's rock is irrelevant if Suzy's rock gets to the bottle first and shatters it.  I suggest the blindingly obvious answer:  it just got there {\em too late}.  By the time Billy's rock reached the bottle, the damage was already done, the bottle lay in pieces, and there was nothing left to shatter.  In other words, one simply cannot cause what is already the case.  My notion of relevance will comprise just this: whatever happened {\em after} the effect  is irrelevant, and whatever happened  {\em on the way to} the effect is counted as relevant. Of course, this is not yet a complete dichotomy, since it does not rule on the status of those events that did not happen at all.  Recall that if some CP-law does not happen in a particular branch, this means that, somewhere along the way, its precondition must have become impossible.  Whether an event that did not happen is considered relevant will depend on when its precondition became impossible: if this was {\em before} the effect arose, then it is relevant, otherwise not. This leads to the following definition.

\begin{definition}[Actual causation in a complete information setting] Let $b = (s_0,\ldots,s_n)$ be a branch of an execution model of a theory $T$.  Let $C$ and $E$ be two atoms that both hold in the final state $s_n$ of $b$, i.e., $\{C,E\} \subseteq \ci(s_n)$. $C$ is an {\em actual cause} of $E$ in branch $b$ if $\pi^{I'}_{T''}(E) = 0$ with $I' = \ci(s_0) \setminus\{C\}$ and  $T'' = ((T')^b)^{\lnot C}$, where $T'$ is constructed as follows.  If $j$ is the smallest $k$ for which $E \in \ci(s_k)$, then $T' = \{\ce(s_i) \mid 0\leq i< j \} \cup \{ r \in T \mid \cu(j-1) \models \lnot body(r)\}$.  In words, $C$ is an actual cause of $E$ if there is a counterfactual dependency from $C$ to $E$, according to the theory $T '$ that consists of both those events that happened before $E$ was caused, and those events that had already become impossible by then. \label{def:actc}
\end{definition}

It is quite easy to check whether this definition is satisfied: you look at the given branch, find the place where $E$ first appeared, discard all events that had not yet happened then but still were possible, and check whether the remaining theory exhibits a counterfactual dependency between $C$ and $E$ or not.  To illustrate, consider again the leftmost branch $(s_0,s_1,s_2)$ of the left execution model for the Billy and Suzy example.  The bottle breaks in node $s_1$, i.e, $Shatters \in \ci(s_1) \setminus \ci(s_0)$.  Therefore, $T' = \{\ce(s_0)\} = \{
(Shatters : 0.9) \leftarrow Throws(Suzy)\}$ and $(T')^b = \{ Shatters \leftarrow Throws(Suzy)\}$.  According to $(T')^b$, there now is indeed a counterfactual dependency from $Throws(Suzy)$ to $Shatters$, so the first is an actual cause of the second.  
As this example illustrates, it is important that a branch $(s_0,\ldots,s_n)$ of an execution model not only records the successive states $\ci(s_i)$ of the domain, but also the events $\ce(s_i)$ that caused each of the state transitions.

Recall that the HP setting offers no mathematical objects that correspond to a complete story about what happened, so their definition is always just given the final outcome in the form of an assignment of values to the RVs.  In this case, we cannot always say with certainty whether some potential cause actually caused an effect or not.  Indeed, if we get only the final interpretation $\ci(s_n)$ instead of the full branch $(s_o,\ldots,s_n)$, then the best we can do is this:

\begin{definition}[Actual causation in a partial information setting]
\label{partial} Let $T$ be a CP-theory and $I$ an interpretation for its vocabulary.  Let $B(I)$ be the set of all branches of all execution models of $T$ that end in a state $s$ for which $\ci(s) = I$.  If $C$ actually causes $E$ in {\em at least one} branch $b \in B(I)$, we say that $C$ is a {\em possible} actual cause for $E$.  If $C$ actually causes $E$ in {\em all} branches $b\in B(I)$, we say that $C$ is a {\em certain} actual cause for $E$.
\end{definition} 

If, in the bottle breaking example, we are only told that eventually $Throws(Suzy)$, $Throws(Billy)$ and $Shatters$ all hold, we find ourselves faced with precisely the same problem as HP's first structural model: all that we can say is that both are possible actual causes, but neither is a certain actual cause. This is typical for redundant causation patterns, and fits well with intuition here: without knowledge about the order in which events happened, we cannot say which of the redundant causes actually ``got there first''.

So far, we have only considered actual causation as it applies to atoms causing atoms.  Often, it is also interesting to wonder which omissions contributed to an effect (``did the doctor's failure to treat the patient cause his death?'') or why some effect was in fact not caused (``did the doctor's treatment prevent the patient's death?'').  Extending the framework to also address such questions is easy enough:
\begin{itemize}
\item To extend our definition of actual causation to allow also literals $\lnot E$ to act as effects, we need to specify when such a $\lnot E$ ``happens'' for the first time, such that we may discard all later events when making counterfactual judgments to determine what caused $\lnot E$.  The obvious cut-off point is when $E$ no longer belongs to the overestimate $\mathcal{U}(s)$. 
\item To also allow literals $\lnot C$ to act as causes, we need to define precisely how we will check the counterfactual dependency in this case.  To assume that $\lnot C$ was not the case, we need to assume that $C$ has somehow occurred, which we can do formally  by just adding a new CP-law ``$C\leftarrow$'' that always causes $C$.
\end{itemize}
Due to space restrictions,  formal details are left to the reader.

\section{Examples}\label{sec:ex}

There is a large literature about actual causation, with many examples, counterexamples, and counter-counterexamples. While e.g.~\citeN{stonesoup} have argued that the importance of such small examples should not be exaggerated,  it nevertheless remains useful to check that my approach behaves sensibly for them.  Due to space restrictions, I will limit myself to those examples that most clearly illustrate the difference between my approach and the HP account.

It is common practice in research on actual causation to formulate examples in terms of neuron diagrams. A neuron can be in one of two states, one is the default ``off'' state and the other is the deviant ``on'' state in which the neuron ``fires'' or ``is active''.  Different kinds of links between two nodes define how the state of one affects the other.  For instance, in the following figure, $E$ fires if and only if $B$ fires, and $B$ fires if at least one of $A$ or $C$ fires. 
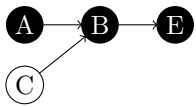
\begin{center}
\begin{pgfpicture}
\pgfsetyvec{\pgfxy(0,0.8)}
\pgfnodecircle{a}[fill]{\pgfxy(0,1)}{0.25cm}
\pgfnodecircle{b}[fill]{\pgfxy(1,1)}{0.25cm}
\pgfnodecircle{e}[fill]{\pgfxy(2,1)}{0.25cm}
\pgfnodecircle{c}[stroke]{\pgfxy(0,0)}{0.25cm}
\pgfsetendarrow{\pgfarrowto}
\pgfnodeconnline{a}{b}
\pgfnodeconnline{b}{e}
\pgfnodeconnline{c}{b}
\color{white}
\pgfputat{\pgfnodecenter{a}}{\pgfbox[center,center]{A}}
\pgfputat{\pgfnodecenter{e}}{\pgfbox[center,center]{E}}
\pgfputat{\pgfnodecenter{b}}{\pgfbox[center,center]{B}}
\color{black}
\pgfputat{\pgfnodecenter{c}}{\pgfbox[center,center]{C}}
\end{pgfpicture}
\end{center}
Neuron diagrams typically record not only this causal structure, but also the state of the neurons.  In the figure above, nodes that are ``on'' are represented by full circles and nodes that are ``off'' are shown as empty circles.  So, $A$, $B$ and $E$ all fire, whereas $C$ does not.  In the language that we have developed so far, a neuron diagram therefore places us in the partial information setting of Definition \ref{partial}: we are given a causal model of a domain together with the final state that has been reached, but are not told precisely how this state has come about.

\citeN{hall07} shows a number of counterexamples to HP, and introduces an alternative account, which he formalizes for neuron diagrams only. One of his counterexamples concerns the following two diagrams:
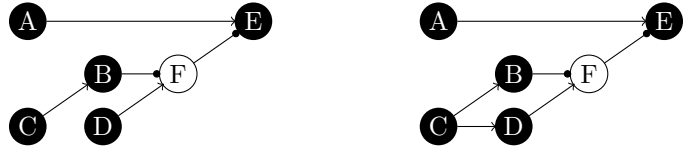
\begin{center}
\hspace{\stretch{1}}
\parbox[c]{3.5cm}{
\begin{pgfpicture}
\pgfsetyvec{\pgfxy(0,0.7)}
\pgfnodecircle{a}[fill]{\pgfxy(0,2)}{0.25cm}
\pgfnodecircle{b}[fill]{\pgfxy(1,1)}{0.25cm}
\pgfnodecircle{e}[fill]{\pgfxy(3,2)}{0.25cm}
\pgfnodecircle{c}[fill]{\pgfxy(0,0)}{0.25cm}
\pgfnodecircle{d}[fill]{\pgfxy(1,0)}{0.25cm}
\pgfnodecircle{f}[stroke]{\pgfxy(2,1)}{0.25cm}
\pgfsetendarrow{\pgfarrowto}
\pgfnodeconnline{a}{e}
\pgfnodeconnline{c}{b}
\pgfnodeconnline{d}{f}
\pgfsetendarrow{\pgfarrowdot}
\pgfnodeconnline{f}{e}
\pgfnodeconnline{b}{f}
\color{white}
\pgfputat{\pgfnodecenter{a}}{\pgfbox[center,center]{A}}
\pgfputat{\pgfnodecenter{b}}{\pgfbox[center,center]{B}}
\pgfputat{\pgfnodecenter{c}}{\pgfbox[center,center]{C}}
\pgfputat{\pgfnodecenter{d}}{\pgfbox[center,center]{D}}
\pgfputat{\pgfnodecenter{e}}{\pgfbox[center,center]{E}}
\color{black}
\pgfputat{\pgfnodecenter{f}}{\pgfbox[center,center]{F}}
\end{pgfpicture}}
\hspace{\stretch{1}}
\parbox[c]{3.5cm}{
\begin{pgfpicture}
\pgfsetyvec{\pgfxy(0,0.7)}
\pgfnodecircle{a}[fill]{\pgfxy(0,2)}{0.25cm}
\pgfnodecircle{b}[fill]{\pgfxy(1,1)}{0.25cm}
\pgfnodecircle{e}[fill]{\pgfxy(3,2)}{0.25cm}
\pgfnodecircle{c}[fill]{\pgfxy(0,0)}{0.25cm}
\pgfnodecircle{d}[fill]{\pgfxy(1,0)}{0.25cm}
\pgfnodecircle{f}[stroke]{\pgfxy(2,1)}{0.25cm}
\pgfsetendarrow{\pgfarrowto}
\pgfnodeconnline{a}{e}
\pgfnodeconnline{c}{b}
\pgfnodeconnline{c}{d}
\pgfnodeconnline{d}{f}
\pgfsetendarrow{\pgfarrowdot}
\pgfnodeconnline{f}{e}
\pgfnodeconnline{b}{f}
\color{white}
\pgfputat{\pgfnodecenter{a}}{\pgfbox[center,center]{A}}
\pgfputat{\pgfnodecenter{b}}{\pgfbox[center,center]{B}}
\pgfputat{\pgfnodecenter{c}}{\pgfbox[center,center]{C}}
\pgfputat{\pgfnodecenter{d}}{\pgfbox[center,center]{D}}
\pgfputat{\pgfnodecenter{e}}{\pgfbox[center,center]{E}}
\color{black}
\pgfputat{\pgfnodecenter{f}}{\pgfbox[center,center]{F}}
\end{pgfpicture}}
\hspace*{\stretch{1}}
\end{center}
In both diagrams, the edges from $B$ to $F$ and from $F$ to $E$ are {\em blocking} edges: if $B$ fires, then $F$ will never fire, regardless of its other incoming edges.  In the left diagram, both $A$ and $C$ cause $E$: $A$ causes it directly and $C$ causes it by stopping $D$ from preventing $E$.  In the right diagram, however, $C$ also causes the very ``threat'' to $E$ that it prevents.  Therefore, Hall argues, in this diagram it should not be counted as a cause for $E$. 

The HP account correctly handles the left diagram, but fails for the right one,  since taking $\vec{X} = \{C\}$ and $\vec{W} = \{D\}$ allows us to create the context $D = \true$ in which there is a counterfactual dependency from $C$ to $E$. 

To see how my definition fares, here are the obvious CP-logic versions. In the first, $A,C$ and $D$ are all exogenous, while in the second only $A$ and $C$ are. \\
\begin{minipage}{0.5\textwidth}
\begin{align}
\label{afe} E& \leftarrow A\land \lnot F.\\
\label{bdf} F& \leftarrow D\land \lnot B.\\
\label{cb} B& \leftarrow C.
\end{align}
\end{minipage}
\begin{minipage}{0.5\textwidth}
\begin{align}
\label{afe2}E& \leftarrow A\land \lnot F.\\
\label{bdf2} F& \leftarrow D\land \lnot B.\\
\label{cb2}B& \leftarrow C.\\
\label{cd2}D& \leftarrow C.
\end{align}
\end{minipage}

First, consider the left theory.  Here, $E$ can only be caused after  \eqref{afe} has already happened and both \eqref{bdf} and \eqref{cb} have become impossible.  Therefore, all these CP-laws are relevant and we end up having to check whether there is a counterfactual dependency from $C$ to $E$ in the original theory.  Clearly, this is the case, since no tree that starts from a root in which the exogenous predicates $D$ and $A$ are {\true} and $C$ is {\false} can produce $E$.   In the second theory, the event \eqref{cd2} may either happen before $E$ is caused or after.  This means we either have to check for a counterfactual dependency in the theory $\{\eqref{afe2},\eqref{bdf2},\eqref{cb2},\eqref{cd2}\}$ or in $\{\eqref{afe2},\eqref{bdf2},\eqref{cb2}\}$
.   In neither theory we find a counterfactual dependency, so $C$ is correctly judged to certainly not be an actual cause of $E$.
HP's problems with this example are caused by the fact that, lacking an explicitly dynamic semantics, they have to resort to interventions to eliminate irrelevant events from consideration.  As an undesired side-effect, they end up allowing the possibility that $D$ itself is relevant for judging the impact of $C$ on $E$, but the link between $C$ and $D$ is not. 

The following is an example of  {\em bogus prevention} \cite{hiddleston05}, taken from \citeN{hitchcock07}.

\begin{quote} \noindent Assassin is in possession of a lethal poison, but has a last minute change of heart and refrains from putting it in Victim's coffee. Bodyguard puts antidote in the coffee, which would have neutralized the poison had there been any.  Victim drinks the coffee and survives.
\end{quote}
Here, HP, as well as others such as \citeN{hitchcock01}, erroneously designate the bodyguard's unnecessary antidote as an actual cause for Victim's survival.  As I will now show, my account handles this correctly.  Since the example states that Assassin has his $ChangeOfHeart$ before the $Antidote$ is administered, I will not make these exogenous atoms, but instead include them as endogenous atoms that are vacuously caused with some unknown (and irrelevant) probability.
\begin{align}
(Antidote:*) &\leftarrow. \label{anti}\\
(ChangeOfHeart:*)&\leftarrow.\label{coh} \\
Poison &\leftarrow \lnot ChangeOfHeart.\label{pois}\\
Death &\leftarrow Poison \land \lnot Antidote.\label{dth}
\end{align}
The example now tells the story that first event \eqref{coh} happens, which is then followed by \eqref{anti}.  However, as soon as \eqref{coh} happens,  both $Poison$ and $Death$ become impossible, so  \eqref{anti} is considered irrelevant in the actual course of events and will not be part of the theory in which we check for a counterfactual dependency.  Hence, preventing $Antidote$ in this theory has no effect whatsoever upon the Victim's survival, so it is not an actual cause of Victim's survival (but $ChangeOfHeart$ is).  Note that if the antidote were administered before the assassin's change of heart, then it  {\em would} be considered relevant, but still not an actual cause of Victim's survival because then \eqref{pois} would no longer be relevant.  To make the antidote an actual cause of Victim's survival, it would have to be administered after the assassin has {\em failed} to have a change of heart.

\section{Implementation}\label{sec:imp}

A prototype implementation can be downloaded from the following URL:
\begin{center}\tt
\href{http://people.cs.kuleuven.be/~joost.vennekens/actcaus/act.pl}{http://people.cs.kuleuven.be/$\sim$joost.vennekens/actcaus/act.pl}
\end{center}
This small program computes whether an atom is a possible/certain actual cause for an effect in the partial information setting, or an actual cause in the complete information setting.  It was written in SWI-prolog, but should also run in Sicstus or YAP.  Currently, it only handles ground theories without disjunction in rule bodies.

In the partial information setting,  this prototype performs a simple backtracking search over all branches that might generate the given observations. Obviously, this is not an approach that would scale well for larger examples.  The goal of this prototype, however, is just to allow interested people to experiment with my definition, in order to see whether it corresponds to their intuition.  As such, it is not meant to handle problems  larger than the examples typically considered in the actual causation literature.  Future work may investigate better algorithms, e.g., by means of an integration into Riguzzi's \citeyear{riguzzi10} query answering algorithm.

\section{Conclusion and related work}\label{sec:con}




This paper has tried to argue that the HP account of actual causation is flawed for two reasons, both of which stem from their choice of structural models as the formal language to express causal relations.  First, structural models fail to make the distinction between default and deviant values, which has been argued by many authors to play a key role in a correct understanding of actual causation.  Second, the static world-view of structural models is ill-suited to handling dynamic concepts, such as the stories that are part of the input to an actual causation problem.  

Since the HP paper first appeared, it has received a great deal of attention among researchers interested in actual causation, and many counterexample and alternative approaches have been presented.  Most of these, such as \citeN{hitchcock07} or \citeN{hall07}, recognize the importance of the deviant/default distinction.  The problems caused by the mismatch between the static formalism of structural equations and the dynamic problem of actual causation have achieved less attention, even though they are recently also pointed out by \citeN{stonesoup}.  Nevertheless, also these more recent approaches still use static formalisms such as neuron diagrams or variants of structural models.  The main point I hope to make in this paper is that for problems that, like actual causation, require reasoning about the way in which a domain evolves, it pays to have a language with a formal semantics that contains mathematical objects that correspond to such evolutions.  

I have tried to illustrate this by defining a notion of actual causation in the context of CP-logic, a probabilistic logic programming language which can be seen as a modular syntactic representation for Shaferian probability trees, which offer precisely the kind of dynamic representation that is perfectly suited for a study of actual causation.  My definition is based on a counterfactual criterion similar to HP's, but is able to leverage the dynamic nature of CP-logic's semantical objects to come up with a very straightforward notion of relevance, namely, it only considers as relevant those events that happened (or became impossible)  {\em before} the effect first arose.  This is much simpler than the relevance criterion of HP, since I do not have to rely on complex manipulations by means of interventions.

While lacking space for an elaborate review of examples from the literature, I have shown that there are  three examples where my definition beats HP:  already for simple examples of redundant causation, it offers a more elegant account due to its ability to distinguish the complete and partial information settings; it is also able to detect fake causes that simply prevent themselves from preventing the effect; and it also handles bogus prevention.   Of course, that is not to say my approach is perfect.  For instance, the railroad switch example from the HP paper cannot  be handled,  because it contains a RV ($Destination$) whose default and deviant values switch in the middle of the story.  I am also offering a prototype implementation of my definition, in the hope that it may help to find further examples where it does not correspond to intuition.  Feedback will be appreciated.

\bibliographystyle{acmtrans}

\end{document}